\documentclass{article}

\PassOptionsToPackage{numbers, compress}{natbib}

\usepackage[preprint]{neurips_2019}

\usepackage[utf8]{inputenc} 
\usepackage[T1]{fontenc}    
\usepackage{hyperref}       
\usepackage{url}            
\usepackage{amsfonts}       
\usepackage{nicefrac}       
\usepackage{microtype}      
\usepackage{bm}
\usepackage{amssymb}
\usepackage{amsmath}
\usepackage{graphicx}
\usepackage{booktabs}       
\usepackage{makecell}
\usepackage{multirow}
\usepackage{footmisc}
\usepackage{float}
\usepackage[affil-it]{authblk}
\date{}
\newcommand{\PreserveBackslash}[1]{\let\temp=\\#1\let\\=\temp}
\newcolumntype{C}[1]{>{\PreserveBackslash\centering}p{#1}}
\newcolumntype{R}[1]{>{\PreserveBackslash\raggedleft}p{#1}}
\newcolumntype{L}[1]{>{\PreserveBackslash\raggedright}p{#1}}

\title{Variational Denoising Network: Toward Blind Noise Modeling and Removal}

\author[1,2]{Zongsheng Yue}
\author[2]{Hongwei Yong}
\author[1]{Qian Zhao} 
\author[2,3]{Lei Zhang}
\author[1,4,*]{Deyu Meng}
\affil[1]{ School of Mathematics and Statistics, Xi'an Jiaotong University, Shaanxi, China}
\affil[2]{Department of Computing, Hong Kong Polytechnic University, Kowloon, Hong Kong}
\affil[3]{DAMO Academy, Alibaba Group, Shenzhen, China}
\affil[4]{Faculty of Information Technology, The Macau University of Science and Technology, Macau, China}
\affil[*]{Corresponding author: dymeng@mail.xjtu.edu.cn}

\begin{document}

\maketitle

\vspace{-6mm}\begin{abstract}
Blind image denoising is an important yet very challenging problem in computer vision due to the complicated
acquisition process of real images. In this work we propose a new variational inference method, which
integrates both noise estimation and image denoising into a unique Bayesian framework, for blind image denoising.
Specifically, an approximate posterior, parameterized by deep neural networks, is presented by taking the intrinsic
clean image and noise variances as latent variables conditioned on the input noisy image. This posterior provides
explicit parametric forms for all its involved hyper-parameters, and thus can be easily implemented for blind
image denoising with automatic noise estimation for the test noisy image. On one hand, as other data-driven deep
learning methods, our method, namely variational denoising network (VDN), can perform denoising efficiently due to
its explicit form of posterior expression. On the other hand, VDN inherits the advantages of traditional model-driven
approaches, especially the good generalization capability of generative models. VDN has good interpretability
and can be flexibly utilized to estimate and remove complicated non-i.i.d. noise collected in real scenarios.
Comprehensive experiments are performed to substantiate the superiority of our method in blind image denoising.
\end{abstract}

\vspace{-3mm}\section{Introduction}\vspace{-3mm}
Image denoising is an important research topic in computer vision, aiming at
recovering the underlying clean image from an observed noisy one. The noise contained in a real noisy image is
generally accumulated from multiple different sources, e.g., capturing instruments, data transmission media,
image quantization, etc.~\cite{tsin2001statistical}. Such complicated
generation process makes it fairly difficult to access the noise information accurately and
recover the underlying clean image from the noisy one. This constitutes the main aim of blind image denoising.

There are two main categories of image denoising methods. Most classical methods belong to the first category, mainly focusing on constructing a rational maximum a posteriori (MAP) model, involving
the fidelity (loss) and regularization terms, from a Bayesian perspective~\cite{bishop2006pattern}.
An understanding for data generation mechanism is required for designing a rational MAP objective, especially
better image priors like sparsity~\cite{aharon2006k},
low-rankness~\cite{gu2014weighted,zhu2016noise,Xu_2018_ECCV}, and non-local
similarity~\cite{buades2005non,maggioni2013nonlocal}.
These methods are superior mainly in their interpretability naturally led by the Bayesian framework.
They, however, still exist critical limitations due to their assumptions on both
image prior and noise (generally i.i.d. Gaussian), possibly deviating from real
spatially variant (i.e.,non-i.i.d.) noise,
and their relatively low implementation speed
since the algorithm needs to be re-implemented for any
new coming image. Recently, deep learning approaches represent a new trend along this research line.
The main idea is to firstly collect large amount of noisy-clean image pairs
and then train a deep neural network denoiser
on these training data in an end-to-end learning manner. This approach is
especially superior in its effective accumulation of knowledge from large datasets and
fast denoising speed for test images.
They, however,  are easy to overfit
to the training data with certain noisy types, and still could not be generalized well on test
images with unknown but complicated noises.

Thus, blind image denoising especially for real images is still a challenging task, since the real noise
distribution is difficult to be pre-known (for model-driven MAP approaches)
and hard to be comprehensively simulated by training data (for data-driven deep learning approaches).

Against this issue, this paper proposes a new variational inference method, aiming at directly inferring both
the underlying clean image and the noise distribution from an observed noisy image in a unique Bayesian framework.
Specifically, an approximate posterior is presented by taking the intrinsic clean image and noise variances as
latent variables conditioned on the input noisy image. This posterior provides explicit parametric forms for all
its involved hyper-parameters, and thus can be efficiently implemented for blind image denoising with automatic
noise estimation for test noisy images.

In summary, this paper mainly makes following contributions: 1) The proposed method is capable of simultaneously
implementing both noise estimation and blind image denoising tasks in a unique Bayesian framework.
The noise distribution is modeled as a general non-i.i.d. configurations with spatial relevance across the image,
which evidently better complies with the heterogeneous real noise beyond the conventional i.i.d. noise assumption.
2) Succeeded from the fine generalization capability of the generative model, the proposed method is verified to
be able to effectively estimate and remove complicated non-i.i.d. noises in test images even though such noise types
have never appeared in training data, as clearly shown in Fig.~\ref{fig:butterfly_case2}. 3) The proposed method is
a generative approach outputted a complete distribution revealing how the noisy image is generated. This not only
makes the result with more comprehensive interpretability beyond traditional methods purely aiming at obtaining
a clean image, but also naturally leads to a learnable likelihood (fidelity) term according to the data-self. 4) The
most commonly utilized deep learning paradigm, i.e., taking MSE as loss function and training on large noisy-clean
image pairs, can be understood as a degenerated form of the proposed generative approach. Their overfitting issue
can then be easily explained under this variational inference perspective: these methods intrinsically put dominant
emphasis on fitting the priors of the latent clean image, while almost neglects the effect of noise variations. This
makes them incline to overfit noise bias on training data and sensitive to the distinct noises in test noisy images.

The paper is organized as follows: Section 2 introduces related work. Sections 3 presents the 
proposed full Bayesion model,
the deep variational inference algorithm, the network architecture and some discussions. Section 4
demonstrates experimental results and the paper is finally concluded.

\vspace{-3mm}\section{Related Work}\vspace{-3mm}
We present a brief review for the two main categories of image denoising methods, i.e., model-driven MAP based methods
and data-driven deep learning based methods.

\textbf{Model-driven MAP based Methods:} Most classical image denoising methods belong to this category, through
designing a MAP model with a fidelity/loss term and a regularization one delivering the pre-known
image prior. Along this line, total variation denoising~\cite{rudin1992nonlinear},
anisotropic diffusion~\cite{perona1990scale} and wavelet coring~\cite{simoncelli1996noise} use the statistical
regularities of images to remove the image noise. Later, the nonlocal similarity prior, meaning many small patches
in a non-local image area possess similar configurations, was widely used in image denoising.
Typical ones include CBM3D~\cite{4271520} and non-local means~\cite{buades2005non}. Some dictionary learning
methods~\cite{gu2014weighted,dong2013nonlocal,Xu_2018_ECCV} and Field-of-Experts (FoE)~\cite{roth2009fields},
also revealing certain prior knowledge of image patches, had also been attempted for the task. Several other
approaches focusing on the fidelity term, which are mainly determined by the noise assumption on data. E.g.,
Mulitscale~\cite{lebrun2015multiscale} assumed the noise of each patch and its similar patches in the same
image to be correlated Gaussian distribution, and LR-MoG~\cite{zhu2016noise}, DP-GMM~\cite{yue2018hyperspectral}
and DDPT~\cite{zhu2017blind} fitted the image noise by using Mixture of Gaussian (MoG) as an approximator for noises.

\textbf{Data-driven Deep Learning based Methods:} Instead of pre-setting image prior, deep learning methods
directly learn a denoiser (formed as a deep neural network) from noisy to clean ones on a large collection
of noisy-clean image pairs. Jain and Seung~\cite{jain2009natural} firstly adopted a five layer convolution neural
network (CNN) for the task. Then some auto-encoder based methods~\cite{xie2012image,agostinelli2013adaptive} were
applied. Meantime, Burger et al.~\cite{burger2012image} achieved the comparable performance with
BM3D using plain multi-layer perceptron (MLP). Zhang et al.~\cite{zhang2017beyond} further proposed the denoising
convolution network (DnCNN) and achieved state-of-the-art performance on Gaussian denoising tasks.
Mao et al.~\cite{mao2016image} proposed a deep fully convolution encoding-decoding network with symmetric
skip connection. Tai et al.~\cite{tai2017memnet} preposed a very deep persistent memory network (MemNet) to
explicitly mine persistent memory through an adaptive learning process. Recently, NLRN~\cite{liu2018non}, N3Net~\cite{plotz2018neural}
and UDNet~\cite{lefkimmiatis2018universal} all embedded the non-local property of image into DNN to 
facilitate the denoising task. In order to boost the flexibility against
spatial variant noise, FFDNet~\cite{zhang2018ffdnet} was proposed by pre-evaluating the noise level and inputting it to
the network together with the noisy image. Guo et al.~\cite{guo2018toward} and
Brooks et al.~\cite{brooks2018unprocessing} both attempted to simulate the generation process of the images in camera.

\vspace{-2mm}\section{Variational Denoising Network for Blind Noise Modeling}\vspace{-2mm}
Given training set $D=\{\bm{y}_j,\bm{x}_j\}_{j=1}^n$, where $\bm{y}_j,\bm{x}_j$ denote the $j^{th}$ training
pair of noisy and the expected clean images, $n$ represents the number of training images,
 our aim is to construct a variational parametric
approximation to the posterior of the latent variables, including the latent clean image and the noise variances,
conditioned on the noisy image. Note that for the noisy image $\bm{y}$, its training pair $\bm{x}$ is
generally a simulated ``clean'' one obtained as the average of many noisy ones taken under similar camera
conditions~\cite{anaya2014renoir,Abdelhamed_2018_CVPR}, and thus is always not the exact latent clean image $\bm{z}$.
This explicit parametric posterior can then be used to directly infer the clean image and noise
distribution from any test noisy image. To this aim, we first need to formulate a rational full Bayesian
model of the problem based on the knowledge delivered by the training image pairs.

\vspace{-2mm}\subsection{Constructing Full Bayesian Model Based on Training Data}\vspace{-2mm}
Denote $\bm{y}=[y_1,\cdots,y_d]^T$ and $\bm{x}=[x_1,\cdots,x_d]^T$ as any training pair in $D$,
where $d$ (width*height) is the size of a training image\footnote{We use $j~(=1,\cdots,n)$ and
$i~(=1,\cdots,d)$ to express the indexes of training data and data dimension, respectively, throughout
the entire paper.}. We can then construct the following model to express the generation process
of the noisy image $\bm{y}$:
\begin{equation}
    y_i \sim \mathcal{N}(y_i|z_i, \sigma_i^2),~ i=1,2,\cdots,d,
    \label{Eq:generate_noisy_im}
\end{equation}
where $\bm{z} \in \mathbb{R}^d$ is the latent clean image underlying $\bm{y}$, $\mathcal{N}(\cdot | \mu, \sigma^2)$ denotes
the Gaussian distribution with mean $\mu$ and variance $\sigma^2$.
Instead of assuming i.i.d. distribution for the noise as
conventional~\cite{mairal2008sparse,dong2013nonlocal,gu2014weighted,Xu_2018_ECCV}, which largely deviates the
spatial variant and signal-depend characteristics of the real noise~\cite{zhang2018ffdnet,brooks2018unprocessing},
we models the noise as a non-i.i.d. and pixel-wise Gaussian distribution in Eq.~\eqref{Eq:generate_noisy_im}.

The simulated ``clean'' image $\bm{x}$ evidently provides a strong prior to the latent variable $\bm{z}$. Accordingly
we impose the following conjugate Gaussian prior on $\bm{z}$:
\begin{equation}
    z_i \sim \mathcal{N}(z_i|x_i, \varepsilon_0^2), ~i=1,2,\cdots,d,
    \label{Eq:prior-z}
\end{equation}
where $\varepsilon_0$ is a hyper-parameter and can be easily set as a small value.

Besides, for $\bm{\sigma}^2=\{\sigma^2_1, \sigma_2^2, \cdots, \sigma_d^2\}$, we also introduce a rational
conjugate prior as follows:
\begin{equation}
    \sigma_i^2 \sim \text{IG}\left(\sigma_i^2|\frac{p^2}{2}-1, \frac{p^2\xi_i}{2}\right), ~ i=1,2,\cdots, d,
    \label{Eq:prior-sigma}
\end{equation}
where $\text{IG}(\cdot|\alpha, \beta)$ is the inverse Gamma distribution with parameter $\alpha$ and $\beta$,
$\bm{\xi}=\mathcal{G}\left( (\hat{\bm{y}}-\hat{\bm{x}})^2;p \right)$ represents the filtering output of the
variance map $(\hat{\bm{y}}-\hat{\bm{x}})^2$ by a Gaussian filter with $p \times p$ window,
and $\hat{\bm{y}}$, $\hat{\bm{x}} \in \mathbb{R}^{h\times w}$ are the matrix (image) forms of $\bm{y}$,
$\bm{x} \in \mathbb{R}^d$, respectively. Note that the mode of above
IG distribution is $\xi_i$~\cite{bishop2006pattern,yong2017robust}, which is a approximate evaluation of $\sigma_i^2$ in $p \times p$ window.

Combining Eqs.~\eqref{Eq:generate_noisy_im}-\eqref{Eq:prior-sigma}, a full Bayesian model for the problem can be obtained.
The goal then turns to
infer the posterior of latent variables $\bm{z}$ and
$\bm{\sigma}^2$ from noisy image $\bm{y}$, i.e., $p(\bm{z}, \bm{\sigma}^2|\bm{y})$.

\begin{figure}[t]
    \centering
    \includegraphics[scale=0.38]{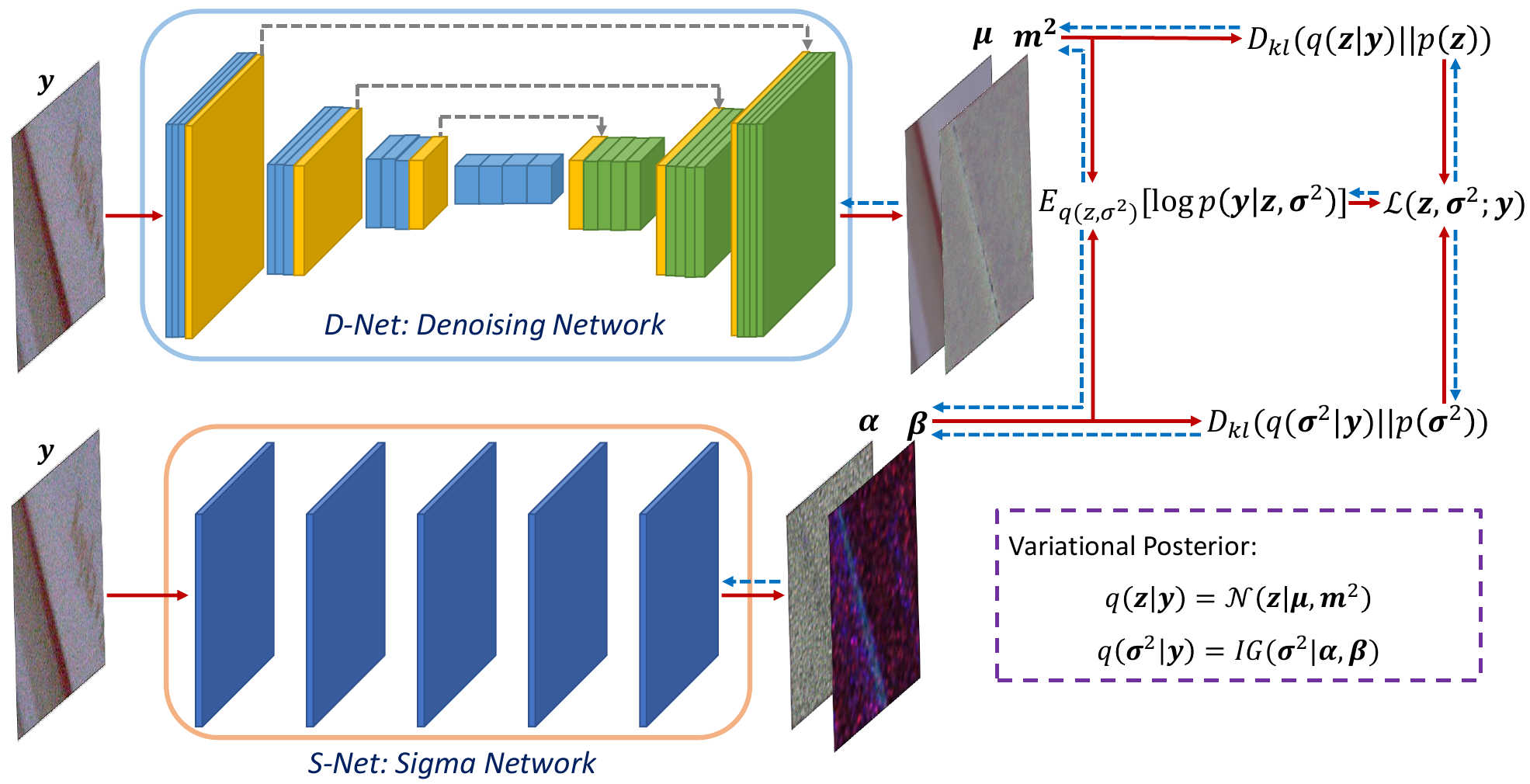}
    \vspace{-3mm}
    \caption{\small{The architecture of the proposed deep variational inference network.
    The red solid lines denote the forward process, and the blue dotted lines mark the gradient
    flow direction in the BP algorithm.}}
    \label{fig:net-loss}\vspace{-4mm}
\end{figure}

\vspace{-2mm}\subsection{Variational Form of Posterior}\vspace{-2mm}
We first construct a variational distribution $q(\bm{z}, \bm{\sigma}^2|\bm{y})$ to approximate the
posterior $p(\bm{z}, \bm{\sigma}^2|\bm{y})$ led by Eqs.~\eqref{Eq:generate_noisy_im}-\eqref{Eq:prior-sigma}. Similar to
the commonly used mean-field variation inference techniques, we assume
conditional independence between variables $\bm{z}$ and $\bm{\sigma}^2$, i.e.,
\begin{equation}
    q(\bm{z}, \bm{\sigma}^2|\bm{y})=q(\bm{z}|\bm{y})q(\bm{\sigma}^2|\bm{y}).
    \label{Eq:posterior-factor-asumption}
\end{equation}
Based on the conjugate priors in Eqs. \eqref{Eq:prior-z} and \eqref{Eq:prior-sigma}, it is natural to
formulate variational posterior forms of $\bm{z}$ and $\bm{\sigma}^2$ as follows:
\begin{align}
    q(\bm{z}|\bm{y}) = \prod_i^d \mathcal{N}(z_i|\mu_i(\bm{y};W_{D}), m_i^2(\bm{y};W_{D})), \
    q(\bm{\sigma}^2|\bm{y}) = \prod_i^d \text{IG}(\sigma_i^2|\alpha_i(\bm{y};W_{S}), \beta_i(\bm{y};W_{S})),
    \label{Eq:distribution_z_sigma}
\end{align}
where $\mu_i(\bm{y};W_{D})$ and $m_i^2(\bm{y};W_{D})$ are designed as the prediction functions for getting
posterior parameters of latent variable $\bm{z}$ directly from $\bm{y}$. The function is represented as a network,
called denoising network or \textit{D-Net}, with parameters $W_{D}$. Similarly, $\alpha_i(\bm{y};W_{S})$ and
$\beta_i(\bm{y};W_{S})$ denote the prediction functions for evaluating posterior parameters of
$\bm{\sigma}^2$ from $\bm{y}$, where $W_{S}$ represents the parameters of the network, called Sigma network or \textit{S-Net}.
The aforementioned is illustrated in Fig. \ref{fig:net-loss}. Our aim is then to optimize
these network parameters $W_{D}$ and $W_{S}$ so as to get the explicit functions for predicting clean image
$\bm{z}$ as well as noise knowledge $\bm{\sigma}^2$ from any test noisy image $\bm{y}$. A rational objective
function with respect to $W_{D}$ and $W_{S}$ is thus necessary
to train both the networks.

Note that the network parameters $W_{D}$ and $W_{S}$ are shared by posteriors calculated on all training data, and
thus if we train them on the entire training set, the method is expected to induce the general statistical inference
insight from noisy image to its underlying clean image and noise level.

\vspace{-2mm}\subsection{Variational Lower Bound of Marginal Data Likelihood}\vspace{-2mm}
For notation convenience, we simply write $\mu_i(\bm{y};W_{D})$, $m_i^2(\bm{y};W_{D})$,
$\alpha_i(\bm{y};W_{S})$, $\beta_i(\bm{y};W_{S})$ as $\mu_i$, $m_i^2$, $\alpha_i$, $\beta_i$ in
the following calculations. For any noisy image $\bm{y}$ and its simulated ``clean'' image $\bm{x}$ in the
training set, we can decompose its marginal likelihood as the following form~\cite{blei2006variational}:
\begin{equation}
    \log p(\bm{y}) = \mathcal{L}(\bm{z},\bm{\sigma}^2;\bm{y})
                            + D_{KL}\left(q(\bm{z},\bm{\sigma}^2|\bm{y})||p(\bm{z},\bm{\sigma}^2|\bm{y}) \right),
    \label{Eq:lower_decomposition}
\end{equation}
where
\begin{equation}
    \mathcal{L}(\bm{z},\bm{\sigma}^2;\bm{y}) = E_{q(\bm{z},\bm{\sigma}^2|\bm{y})}
    \left[\log p(\bm{y}|\bm{z},\bm{\sigma}^2)p(\bm{z})p(\bm{\sigma}^2)-\log q(\bm{z},\bm{\sigma}^2|\bm{y})  \right],
    \label{Eq:lower_bound}
\end{equation}
Here $E_{p(x)}[f(x)]$ represents the exception of $f(x)$ w.r.t. stochastic variable $x$ with probability density
function $p(x)$. The second term of Eq.~\eqref{Eq:lower_decomposition} is a KL divergence between the variational
approximate posterior $q(\bm{z},\bm{\sigma}^2|\bm{y})$ and the true posterior $p(\bm{z},\bm{\sigma}^2|\bm{y})$ with
non-negative value. Thus the first term $\bm{\mathcal{L}}(\bm{z},\bm{\sigma}^2;\bm{y})$ constitutes a
\textit{variational lower bound} on the logarithm of marginal likelihood $p(\bm{y})$, i.e.,
\begin{equation}
    \log p(\bm{y}) \ge \mathcal{L} (\bm{z},\bm{\sigma}^2;\bm{y}).
    \label{Eq:nonequal_bound}
\end{equation}
According to Eqs.~\eqref{Eq:posterior-factor-asumption}, \eqref{Eq:distribution_z_sigma} and \eqref{Eq:lower_bound}, the
lower bound can then be rewritten as:
\begin{equation}
    \mathcal{L}(\bm{z},\bm{\sigma}^2;\bm{y})=E_{q(\bm{z},\bm{\sigma}^2|\bm{y})}
    \left[ \log p(\bm{y}|\bm{z},\bm{\sigma}^2) \right]
    - D_{KL}\left(q(\bm{z}|\bm{y}) || p(\bm{z}) \right)
    - D_{KL} \left(q(\bm{\sigma}^2|\bm{y}) || p(\bm{\sigma^2}) \right).
    \label{Eq:lower_bound_factor}
\end{equation}
It's pleased that all the three terms in Eq~\eqref{Eq:lower_bound_factor} can be integrated
analytically as follows:
\begin{footnotesize}
\begin{equation}
    E_{q(\bm{z},\bm{\sigma}^2|\bm{y})}\left[ \log p(\bm{y}|\bm{z},\bm{\sigma}^2) \right]
    = \sum_{i=1}^d \Big\{-\frac{1}{2}\log 2\pi - \frac{1}{2}(\log \beta_i - \psi(\alpha_i))
    -\frac{\alpha_i}{2\beta_i}\left[ (y_i-\mu_i)^2 + m_i^2\right]\Big\},
    \label{Eq:lower_bound_likeli}
\end{equation}
\begin{equation}
    D_{KL}\left(q(\bm{z}|\bm{y}) || p(\bm{z}) \right) = \sum_{i=1}^d \Big\{
        \frac{(\mu_i-x_i)^2}{2\varepsilon_0^2}
        + \frac{1}{2}\left[\frac{m_i^2}{\varepsilon_0^2}-\log \frac{m_i^2}{\varepsilon_0^2} -1 \right]
    \Big\},
    \label{Eq:lower_bound_group_kl1} 
\end{equation}
\begin{align}
    D_{KL} \left(q(\bm{\sigma}^2|\bm{y}) || p(\bm{\sigma^2}) \right) &= \sum_{i=1}^d \bigg\{
        \left(\alpha_i-\frac{p^2}{2}+1\right)\psi(\alpha_i)
        + \left[ \log \Gamma \left(\frac{p^2}{2}-1\right) - \log \Gamma (\alpha_i)\right] \notag \\
        &\mathrel{\phantom{=}} \hspace{1.5cm}+\left(\frac{p^2}{2}-1\right)\left(\log \beta_i - \log \frac{p^2\xi_i}{2}\right)
        +\alpha_i \left(\frac{p^2\xi_i}{2\beta_i}-1\right)\bigg\},
        \label{Eq:lower_bound_group_kl2}
\end{align}
\end{footnotesize}
where $\psi(\cdot)$ denotes the digamma function. Calculation details are listed in supplementary material.

We can then easily get the expected objective function (i.e., a negtive lower bound of the marginal likelihood on
entire training set) for optimizing the network parameters of \textit{D-Net} and \textit{S-Net} as follows:
\begin{equation}
\min_{W_D,W_S} -\sum_{j=1}^n \mathcal{L}(\bm{z}_j,\bm{\sigma}_j^2;\bm{y}_j). \label{Eq:Objective}
\end{equation}

\vspace{-4mm}\subsection{Network Learning} \label{sec:inference}\vspace{-2mm}
As aforementioned, we use \textit{D-Net} and \textit{S-Net} together to infer
the variational parameters $\bm{\mu}$, $\bm{m}^2$ and $\bm{\alpha}$, $\bm{\beta}$ from the input
noisy image $\bm{y}$, respectively, as shown in Fig.~\ref{fig:net-loss}. It is critical to consider how to
calculate derivatives of this objective with respect to $W_D,W_S$ involved in $\bm{\mu}$, $\bm{m}^2$,
$\bm{\alpha}$ and $\bm{\beta}$ to facilitate an easy use of stochastic gradient varitional inference. Fortunately,
different from other related variational inference techniques
like VAE \cite{kingma2013auto}, all three terms of Eqs.~\eqref{Eq:lower_bound_likeli}-\eqref{Eq:lower_bound_group_kl2}
in the lower bound Eq.~\eqref{Eq:lower_bound_factor} are differentiable and their derivatives can be calculated
analytically without the need of any reparameterization trick, largely reducing the difficulty of network
training.

At the training stage of our method, the network parameters can be easily updated with backpropagation (BP)
algorithm~\cite{goodfellow2016deep} through Eq.~\eqref{Eq:Objective}. The function of each term in this objective can
be intuitively explained: the first term represents the likelihood of the observed noisy images in training set,
and the last two terms control the discrepancy between the variational posterior and the corresponding prior.
During the BP training process, the gradient information from the likelihood term of Eq.~\eqref{Eq:lower_bound_likeli}
is used for updating both the parameters of \textit{D-Net} and \textit{S-Net} simultaneously, implying that the
inference for the latent clean image $\bm{z}$ and $\bm{\sigma}^2$ is guided to be learned from each other.

At the test stage, for any test noisy image, through feeding it into \textit{D-Net}, the final denoising
result can be directly obtained by $\bm{\mu}$. Additionally, through inputting the noisy image to the \textit{S-Net},
the noise distribution knowledge (i.e., $\bm{\sigma}^2$) is easily inferred. Specifically, the noise variance in
each pixel can be directly obtained by using the mode of the inferred inverse Gamma
distribution: $\sigma_i^2=\frac{\beta_i}{(\alpha_i+1)}$.

\vspace{-2mm}\subsection{Network Architecture}\vspace{-2mm}
The D-Net in Fig.~\ref{fig:net-loss} takes the noisy image $\bm{y}$ as input to infer the variational
parameters $\bm{\mu}$ and $\bm{m}^2$ in $q(\bm{z}|\bm{y})$ of Eq.~\eqref{Eq:distribution_z_sigma},
and performs the denoising task in the proposed variational inference algorithm. In order to capture
multi-scale information of the image, we use a U-Net~\cite{ronneberger2015u} with depth 4 as the D-Net,
which contains 4 encoder blocks ([\textit{Conv}+\textit{ReLU}]$\times$2+\textit{Average pooling}), 3 decoder
blocks (\textit{Transpose Conv}+[\textit{Conv}+\textit{ReLU}]$\times$2) and symmetric skip connection under
each scale. For parameter $\bm{\mu}$, the residual learning strategy is adopted as in~\cite{zhang2017beyond}, i.e.,
$\bm{\mu}=\bm{y}+f(\bm{y};W_D)$, where $f( \cdot ;W_D)$ denotes the \textit{D-Net} with parameters $W_D$.
As for the \textit{S-Net}, which takes the noisy image $\bm{y}$ as input and outputs the predicted variational
parameters $\bm{\alpha}$ and $\bm{\beta}$ in $q(\bm{\sigma}^2|\bm{y})$ of Eq~\eqref{Eq:distribution_z_sigma},
we use the DnCNN~\cite{zhang2017beyond} architecture with five layers, and the feature channels of each layer
is set as 64. It should be noted that our proposed method is a general framework, most of the commonly used network
architectures~\cite{zhang2018ffdnet,Ploetz2018,lefkimmiatis2018universal,zhang2018residual} in image restoration can
also be easily substituted.

\vspace{-2mm}\subsection{Some Discussions}\label{sec:discussion}\vspace{-2mm}
It can be seen that the proposed method succeeds advantages of both model-driven MAP and data-driven deep learning methods.
On one hand, our method is a generative approach and possesses fine interpretability to the data generation mechanism;
and on the other hand it conducts an explicit prediction function, facilitating efficient image denoising as well as
noise estimation directly through an input noisy image. Furthermore, beyond current methods, our method can finely
evaluate and remove non-i.i.d. noises embedded in images, and has a good generalization capability to images with
complicated noises, as evaluated in our experiments. This complies with the main requirement of the blind image
denoising task.

If we set the hyper-parameter $\varepsilon_0^2$ in Eq.\eqref{Eq:prior-z} as an extremely small value close to $0$, it is
easy to see that the objective of the proposed method is dominated by the second term of Eq.~\eqref{Eq:lower_bound_likeli},
which makes the objective degenerate as the MSE loss generally used in traditional deep learning
methods (i.e., minimizing $\sum_{j=1}^n||\bm{\mu}(\bm{y}_j;W_{D})-\bm{x}_j||^2$. This provides a new understanding to
explain why they incline to overfit noise bias in training data. The posterior inference process puts dominant
emphasis on fitting priors imposed on the latent clean image, while almost neglects the effect of noise variations. This
naturally leads to its sensitiveness to unseen complicated noises contained in test images.

Very recently, both CBDNet~\cite{guo2018toward} and FFDNet~\cite{zhang2018ffdnet} are presented for the
denoising task by feeding the noisy image integrated with the pre-estimated noise level into the deep network to
make it better generalize to distinct noise types in training stage. Albeit more or less improving
the generalization capability of network, such strategy is still too heuristic and is not easy to interpret
how the input noise level intrinsically influence the final denoising result. Comparatively, our method is
constructed in a sound Bayesian manner to estimate clean image and noise distribution together from the input
noisy image, and its generalization can be easily explained from the perspective of generative model.
\begin{figure}[t]
    \centering\vspace{-3mm}
    \includegraphics[scale=0.54]{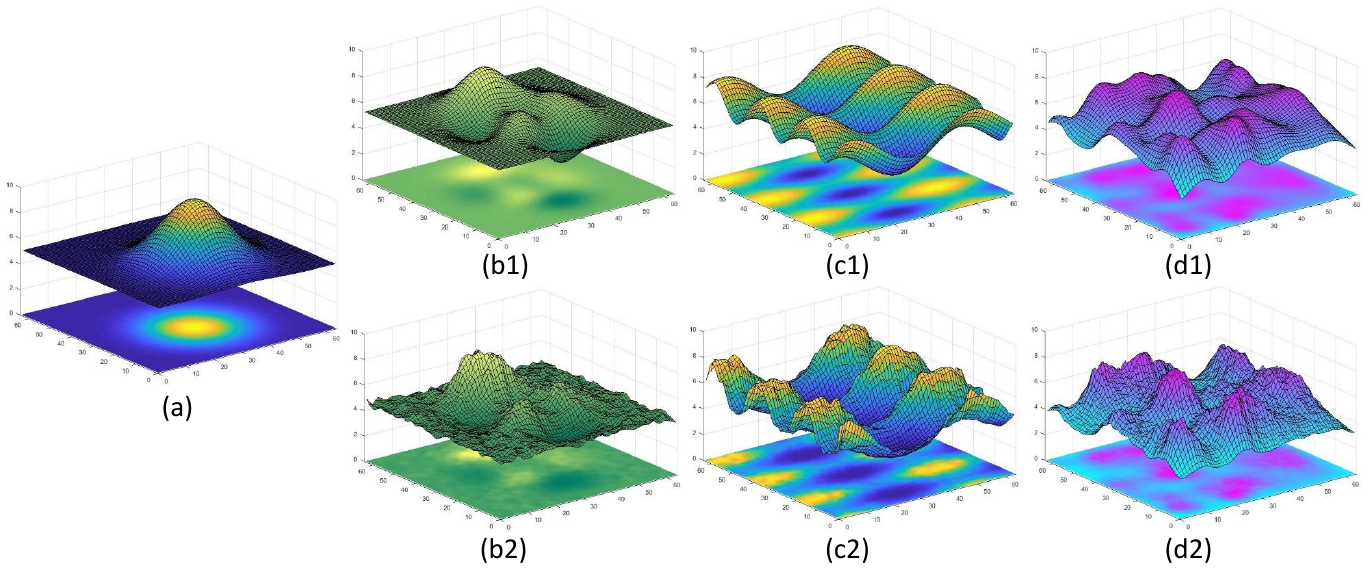}\vspace{-3mm}
    \caption{\small{(a) The spatially variant map $\bm{M}$ for noise generation in training data. (b1)-(d1):
        Three different $\bm{M}$s on testing data in Cases 1-3.
        (b2)-(d2): Correspondingly predicted $\bm{M}$s by our method on the testing data.}}
    \vspace{-2mm}
    \label{fig:sigma_train_test}
\end{figure}
\begin{table}[t]
    \centering
    \caption{\small{The PSNR(dB) results of all competing methods on the three groups of test datasets. The best and second
    best results are highlighted in bold and Italic, respectively.}}\tiny \vspace{-2mm}
    \begin{tabular}{@{}C{1.0cm}@{}|@{}C{1.2cm}@{}|@{}C{1.1cm}@{}@{}C{1.1cm}@{}@{}C{1.1cm}@{}
        @{}C{1.0cm}@{}@{}C{1.0cm}@{}@{}C{1.5cm}@{}@{}C{1.3cm}@{}@{}C{1.3cm}@{}@{}C{1.1cm}@{}@{}C{1.2cm}@{}}
        \Xhline{0.8pt}
        \multirow{2}*{Cases}&\multirow{2}*{Datasets} & \multicolumn{8}{c}{Methods} \\
        \Xcline{3-12}{0.4pt}
        & & \tiny{CBM3D} &\tiny{WNNM}    &\tiny{NCSR}  &\tiny{MLP}  &\tiny{DnCNN-B} &\tiny{MemNet}  &\tiny{FFDNet}
        &\tiny{$\text{FFDNet}_v$} &\tiny{UDNet} &\tiny{VDN} \\
        \Xhline{0.4pt}
        \multirow{3}*{Case 1} & Set5
                    & 27.76 &26.53   &26.62  &27.26  &29.85     & 30.10   &\textit{30.16}  &30.15  &28.13  &\textbf{30.39} \\
        \Xcline{2-12}{0.4pt}                                             
          &  LIVE1  & 26.58 &25.27   &24.96  &25.71  &28.81     & 28.96   &\textit{28.99}  &28.96  &27.19  &\textbf{29.22} \\
        \Xcline{2-12}{0.4pt}                                             
          &  BSD68  & 26.51 &25.13   &24.96  &25.58  &28.73     & 28.74   &\textit{28.78}  &28.77  &27.13  &\textbf{29.02}  \\
        \Xhline{0.4pt}                                                   
        \multirow{3}*{Case 2} & Set5                                     
                    & 26.34 &24.61   &25.76  &25.73  &29.04     & 29.55   &\textit{29.60}  &29.56  &26.01  &\textbf{29.80}  \\
        \Xcline{2-12}{0.4pt}                                             
          &  LIVE1  & 25.18 &23.52   &24.08  &24.31  &28.18     & 28.56   &\textit{28.58}  &28.56  &25.25  &\textbf{28.82} \\
        \Xcline{2-12}{0.4pt}                                             
          &  BSD68  & 25.28 &23.52   &24.27  &24.30  &28.15     & 28.36   &\textit{28.43}  &28.42  &25.13  &\textbf{28.67} \\
        \Xhline{0.4pt}                                                   
        \multirow{3}*{Case 3} & Set5                                     
                    & 27.88 &26.07   &26.84  &26.88  &29.13     & 29.51   &\textit{29.54}  &29.49  &27.54  &\textbf{29.74} \\
        \Xcline{2-12}{0.4pt}                                             
        &  LIVE1    & 26.50 &24.67   &24.96  &25.26  &28.17     & 28.37   &\textit{28.39}  &28.38  &26.48  &\textbf{28.65}   \\
        \Xcline{2-12}{0.4pt}                                             
        &  BSD68    & 26.44 &24.60   &24.95  &25.10  &28.11     & 28.20   &\textit{28.22}  &28.20  &26.44  &\textbf{28.46}   \\
        \Xhline{0.8pt}
    \end{tabular}
    \label{tab:psnr_noniid}
    \vspace{-2mm}
\end{table}

\vspace{-2mm}\section{Experimental Results}\vspace{-2mm}
We evaluate the performance of our method on synthetic and real datasets in this section. All experiments
are evaluated in the sRGB space. We briefly denote our method as VDN in the following. The training and
testing codes of our VDN is available at \url{https://github.com/zsyOAOA/VDNet}.

\vspace{-2mm}\subsection{Experimental Setting}\vspace{-2mm}
\textbf{Network training and parameter setting:} The weights of \textit{D-Net} and \textit{S-Net} in our variational
algorithm were initialized according to~\cite{he2015delving}. In each epoch, we randomly crop $N=64\times 5000$ patches
with size $128\times 128$ from the images for training. The Adam algorithm~\cite{Kingma2015} is adopted to
optimize the network parameters through minimizing the proposed negative lower bound objective. The initial learning
rate is set as $2e\text{-}4$ and linearly decayed in half every 10 epochs until to $1e\text{-}6$.
The window size $p$ in Eq.~\eqref{Eq:prior-sigma} is set as 7. The hyper-parameter $\varepsilon_0^2$ is set as
$5e\text{-}5$ and $1e\text{-}6$ in the following synthetic and real-world image denoising experiments, respectively.

\textbf{Comparison methods:} Several state-of-the-art denoising methods are adopted for performance comparison,
including CBM3D~\cite{4271520}, WNNM~\cite{gu2014weighted}, NCSR~\cite{dong2012nonlocally}, MLP~\cite{burger2012image},
DnCNN-B~\cite{zhang2017beyond}, MemNet~\cite{tai2017memnet}, FFDNet~\cite{zhang2018ffdnet}, UDNet~\cite{lefkimmiatis2018universal}
and CBDNet~\cite{guo2018toward}. Note that CBDNet is mainly designed for blind denoising task, and thus we only
compared CBDNet on the real noise removal experiments.

\vspace{-2mm}\subsection{Experiments on Synthetic Non-I.I.D. Gaussian Noise Cases}\vspace{-2mm}
Similar to~\cite{zhang2018ffdnet}, we collected a set of source images to train the network,
including 432 images from BSD~\cite{amfm_pami2011}, 400 images from the validation set of ImageNet~\cite{DengCHI14} and
4744 images from the Waterloo Exploration Database~\cite{ma2017waterloo}. Three commonly used datasets
in image restoration (Set5, LIVE1 and BSD68 in~\cite{kim2016accurate}) were adopted as test datasets to
evaluate the performance of different methods.
In order to evaluate the effectiveness and robustness of VDN under the non-i.i.d. noise configuration,
we simulated the non-i.i.d. Gaussian noise as following,
\begin{equation}
    \bm{n} = \bm{n}^1 \odot \bm{M}, ~ ~n^1_{ij} \sim \mathcal{N}(0,1),
    \label{eq:noise-generation}
\end{equation}
where $\bm{M}$ is a spatially variant map with the same size as the source image.
We totally generated four kinds of $\bm{M}$s as shown in Fig.~\ref{fig:sigma_train_test}.
The first (Fig.~\ref{fig:sigma_train_test} (a)) is used for generating noisy images of training data
and the others (Fig.~\ref{fig:sigma_train_test} (b)-(d)) generating three groups of testing data (denotes as
Cases 1-3). Under this noise generation manner, the noises in training data and testing data are
with evident difference, suitable to verify the robustness and generalization capability of competing
methods.

\begin{figure}[t]
    \centering
    \includegraphics[scale=0.46]{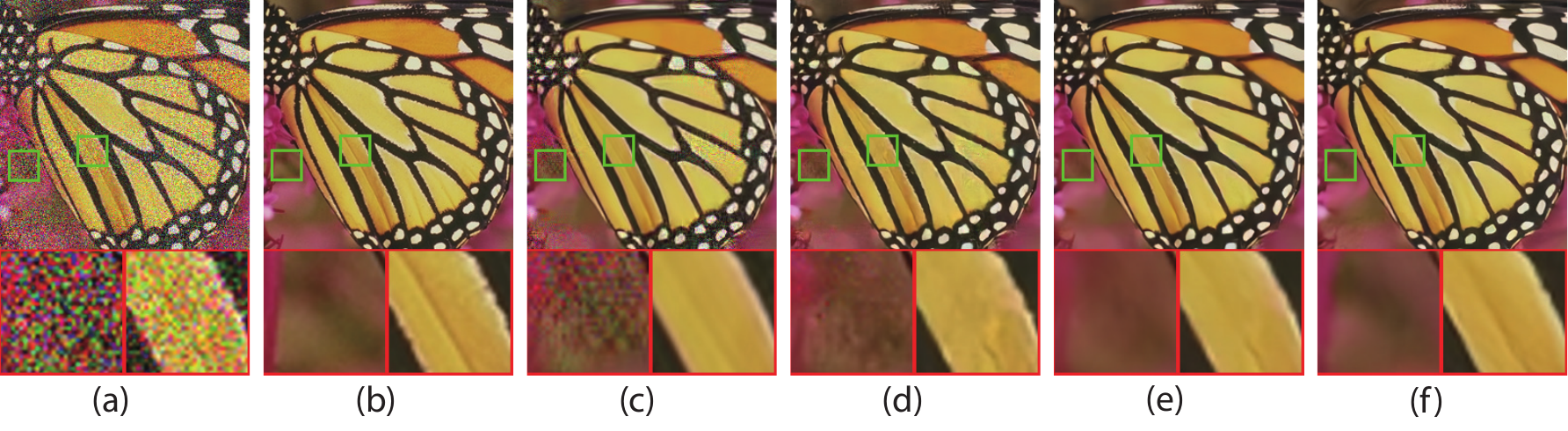}
    \vspace{-3mm}
    \caption{\small{Image denoising results of a typical test image in Case 2. (a) Noisy image, (b) Groundtruth,
    (c) CBM3D (24.63dB), (d) DnCNN-B (27.83dB), (e) FFDNet (28.06dB), (f) VDN (28.32dB).}}
    \label{fig:butterfly_case2} 
\end{figure}

\textbf{Comparson with the State-of-the-art:}
Table~\ref{tab:psnr_noniid} lists the average PSNR results of all competing methods on three groups of testing data.
From Table~\ref{tab:psnr_noniid}, it can be easily observed that: 1) The VDN outperforms
other competing methods in all cases, indicating that VDN is able to handle such
complicated non-i.i.d. noise; 2) VDN surpasses FFDNet about 0.25dB averagely even though
FFDNet depends on the true noise level information instead of automatically inferring noise
distribution as our method;
3) the discriminative methods MLP, DnCNN-B and UDNet seem to evidently overfit on training noise bias;
4) the classical model-driven method CBM3D performs more stably than WNNM and NCSR, possibly due to the latter's
improper i.i.d. Gaussian
noise assumption. Fig.~\ref{fig:butterfly_case2} shows the denoising results of different competing methods on
one typical image in testing set of Case 2, and more denoising results can be found in the supplementary material.
Note that we only display the top four best results from all due to page limitation. It can be seen
that the denoised images by CBM3D and DnCNN-B still contain obvious noise, and FFDNet over-smoothes the image and
loses some edge information, while our proposed VDN removes most of the noise and preserves more details.
\begin{table}[t]
    \centering
    \caption{\small{The PSNR(dB) results of all competing methods on AWGN noise cases of three
    test datasets.}}\tiny\vspace{-2mm}
    \begin{tabular}{@{}C{1.1cm}@{}|@{}C{1.1cm}@{}|@{}C{1.1cm}@{}@{}C{1.1cm}@{}@{}C{1.1cm}@{}
        @{}C{1.0cm}@{}@{}C{1.0cm}@{}@{}C{1.5cm}@{}@{}C{1.3cm}@{}@{}C{1.3cm}@{}@{}C{1.2cm}@{}@{}C{1.2cm}@{}}
        \Xhline{0.8pt}
        \multirow{2}*{Sigma}&\multirow{2}*{Datasets} & \multicolumn{8}{c}{Methods} \\
        \Xcline{3-12}{0.4pt}
                  & & CBM3D  &WNNM    &NCSR  &MLP   &DnCNN-B         &MemNet      &FFDNet            &$\text{FFDNet}_e$ &UDNet &VDN \\
        \Xhline{0.4pt}                                                        
        \multirow{3}*{$\sigma=15$} & Set5                                     
                     &33.42  &32.92   &32.57 &-     &34.04           &34.18       &34.30             &\textit{34.31}    &34.19 &\textbf{34.34} \\
        \Xcline{2-12}{0.4pt}                                                  
           &  LIVE1  &32.85  &31.70   &31.46 &-     & 33.72          &33.84       &\textbf{33.96}    &\textbf{33.96}    &33.74 &33.94 \\
        \Xcline{2-12}{0.4pt}                                                  
           &  BSD68  &32.67  &31.27   &30.84 &-     &\textit{33.87}  &33.76       &33.85             &33.68             &33.76 &\textbf{33.90}  \\
        \Xhline{0.4pt}                                                        
        \multirow{3}*{$\sigma=25$} & Set5                                     
                     &30.92  &30.61   &30.33 &30.55 &31.88           &31.98       &\textit{32.10}    &32.09             &31.82 &\textbf{32.24}  \\
        \Xcline{2-12}{0.4pt}                                                  
           &  LIVE1  &30.05  &29.15   &29.05 &29.16 &31.23           &31.26       &\textit{31.37}    &\textit{31.37}    &31.09 &\textbf{31.50} \\
        \Xcline{2-12}{0.4pt}                                                  
           &  BSD68  &29.83  &28.62   &28.35 &28.93 &\textit{31.22}  &31.17       &31.21             &31.20             &31.02 &\textbf{31.35} \\
        \Xhline{0.4pt}                                                        
        \multirow{3}*{$\sigma=50$} & Set5                                     
                     &28.16  &27.58   &27.20 &27.59 &28.95           &29.10       &\textit{29.25}    &\textit{29.25}    &28.87 &\textbf{29.47} \\
        \Xcline{2-12}{0.4pt}                                                  
           &  LIVE1  &26.98  &26.07   &26.06 &26.12 &27.95           &27.99       &\textit{28.10}    &\textit{28.10}    &27.82 &\textbf{28.36}   \\
        \Xcline{2-12}{0.4pt}                                                  
           &  BSD68  &26.81  &25.86   &25.75 &26.01 &27.91           &27.91       &\textit{27.95}    &\textit{27.95}    &27.76 &\textbf{28.19}   \\
        \Xhline{0.8pt}
    \end{tabular}
    \label{tab:psnr_iidgauss}
\end{table}

Even though our VDN is designed based on the non-i.i.d. noise assumption and trained on the non-i.i.d. noise data,
it also performs well on additive white Gaussian noise (AWGN) removal task.
Table~\ref{tab:psnr_iidgauss} lists the average PSNR results under three noise levels ($\sigma = 15, 25, 50$) of AWGN.
It is easy to see that our method obtains the best or at least comparable performance with the state-of-the-art
method FFDNet.
Combining Table~\ref{tab:psnr_noniid} and Table~\ref{tab:psnr_iidgauss}, it
should be rational to say that our VDN is robust and able to handle a wide range of noise types, due to its better
noise modeling manner.

\textbf{Noise Variance Prediction:} The \textit{S-Net} plays the role of noise modeling and is
able to infer the noise distribution from the noisy image. To verify the fitting capability of \textit{S-Net}, we
provided the $\bm{M}$ predicted by \textit{S-Net} as the input of FFDNet, and the denoising results are listed
in Table~\ref{tab:psnr_noniid} (denoted as $\text{FFDNet}_v$).
It is obvious that FFDNet under the real noise level and $\text{FFDNet}_v$ almost have the same performance,
indicating that the \textit{S-Net} effectively captures proper noise information. 
The predicted noise variance Maps on three groups of testing data are shown in
Fig.~\ref{fig:sigma_train_test}~(b2-d2) for easy observation.
\begin{figure}[t]
    \centering\vspace{-3mm}
    \includegraphics[scale=0.48]{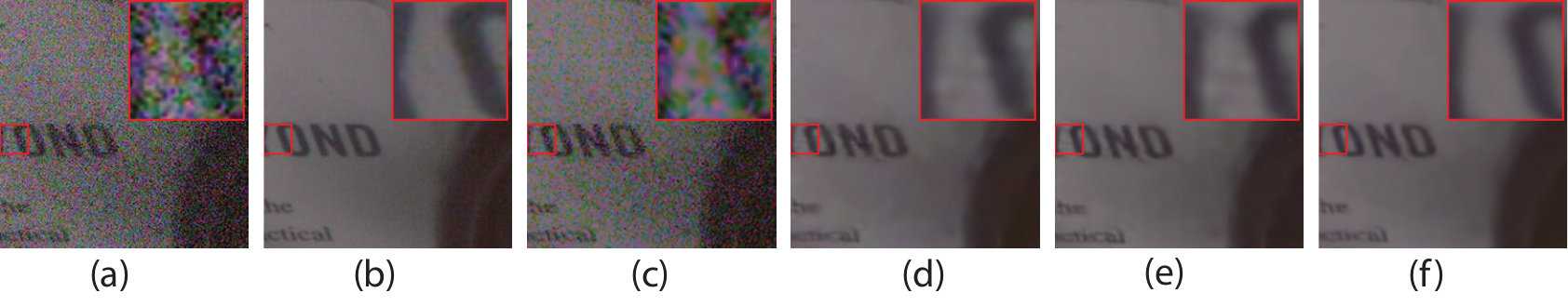}
    \vspace{-2mm}
    \caption{\small{Denoising results on one typical image in the validation set of SIDD. (a) Noisy image,
    (b) Simulated ``clean'' image, (c) WNNM(21.80dB), (d) DnCNN (34.48dB), (e) CBDNet (34.84dB),
    (d) VDN (35.50dB).}}
    \label{fig:SIDD_validation}
\end{figure}
\begin{table}
    \centering
    \caption{The comparison results of different methods on SIDD benchmark and validation dataset.}
    \centering\vspace{-2mm}
    \scriptsize
    \begin{tabular}{@{}C{1.5cm}@{}|@{}C{1.4cm}@{}@{}C{1.4cm}@{}@{}C{1.1cm}@{}@{}C{1.7cm}@{}@{}C{1.4cm}@{}@{}C{1.1cm}@{}|@{}C{1.7cm}@{}@{}C{1.5cm}@{}@{}C{1.1cm}@{}}
        \Xhline{0.8pt}
        Datasets  & \multicolumn{6}{c|}{SIDD Benchmark} & \multicolumn{3}{c}{SIDD Validation} \\
        \Xhline{0.4pt}
        Methods   & CBM3D  & WNNM  & MLP   & DnCNN-B & CBDNet & VDN              & DnCNN-B          & CBDNet & VDN           \\
        \Xhline{0.4pt}
        PSNR      & 25.65  & 25.78 & 24.71 & 23.66   & 33.28  & \textbf{39.26}   & 38.41            &38.68   &\textbf{39.28} \\
        \Xhline{0.4pt}
        SSIM      & 0.685  & 0.809 & 0.641 & 0.583   & 0.868  & \textbf{0.955}   &\textbf{0.909}    &0.901   &\textbf{0.909}        \\
        \Xhline{0.8pt}
    \end{tabular}
    \label{tab:PSNR-SSIM-SIDD}
\end{table}
\begin{table}[!t]
    \centering\vspace{-1mm}
    \caption{\small{The comparison results of all competing methods on DND benchmark dataset.}}\vspace{-2mm}
    \scriptsize
    \begin{tabular}{@{}C{1.5cm}@{}|@{}C{1.6cm}@{}@{}C{1.5cm}@{}@{}C{1.5cm}@{}@{}C{1.3cm}@{}@{}C{1.5cm}@{}@{}C{1.5cm}@{}
        @{}C{1.6cm}@{}@{}C{1.8cm}@{}}
        \Xhline{0.8pt}
        Methods & CBM3D    & WNNM    & NCSR   & MLP    & DnCNN-B          & FFDNet & CBDNet          & VDN \\
        \Xhline{0.4pt}
        PSNR    &34.51     & 34.67   & 34.05  & 34.23  & 37.90            & 37.61  &38.06            & \textbf{39.38}  \\
        \Xhline{0.4pt}     
        SSIM    &0.8507    & 0.8646  & 0.8351 & 0.8331 &0.9430            & 0.9415 &0.9421           & \textbf{0.9518}  \\
        \Xhline{0.8pt}
    \end{tabular}
    \label{tab:PSNR-DND}
    \vspace{-3mm}
\end{table}

\vspace{-4mm}\subsection{Experiments on Real-World Noise}\vspace{-2mm}
In this part, we evaluate the performance of VDN on real blind denoising task, including two banchmark
datasets: DND~\cite{plotz2017benchmarking} and SIDD~\cite{Abdelhamed_2018_CVPR}. DND consists of 50
high-resolution images with realistic noise from 50 scenes taken by 4 consumer cameras.
However, it does not provide any other additional noisy and clean image pairs to train the network.
SIDD~\cite{Abdelhamed_2018_CVPR} is another real-world denoising benchmark, containing $30,000$ real noisy images captured
by 5 cameras under 10 scenes. For each noisy image, it estimates one simulated ``clean'' image through
some statistical methods~\cite{Abdelhamed_2018_CVPR}.
About 80$\%$ ($\sim 24,000$ pairs) of this dataset are provided for training purpose, and the rest as held for benchmark.
And 320 image pairs selected from them are packaged together as a medium version of SIDD, called SIDD Medium
Dataset\footnote{\label{foot:SIDD}https://www.eecs.yorku.ca/~kamel/sidd/index.php},
for fast training of a denoiser.
We employed this medium vesion dataset to train a real-world image denoiser, and test the performance on the two benchmarks.

Table~\ref{tab:PSNR-SSIM-SIDD} lists PSNR results of different methods on
SIDD benchmark\footnote{We employed the function 'compare\_ssim' in scikit-image library to calculate
the SSIM value, which is a little difference with the SIDD official results}.
Note that we only list the results of the competing methods that are available
on the official benchmark website\footref{foot:SIDD}.
It is evident that VDN outperforms other methods.
However, note that neither DnCNN-B nor CBDNet performs well, possibly because they were trained
on the other datasets, whose noise type is different from SIDD. For fair comparison, we retrained DnCNN-B and CBDNet
based on the SIDD dataset. The performance on the SIDD validation
set is also listed in Table~\ref{tab:PSNR-SSIM-SIDD}. Under same training conditions, VDN still outperforms DnCNN-B 0.87 PSNR
and CBDNet 0.60dB PSNR, indicating the effectiveness and significance of our non-i.i.d. noise modeling manner.
For easy visualization, on one typical denoising example, results of the best four competing methods are displayed
in Fig.~\ref{fig:SIDD_validation}

Table~\ref{tab:PSNR-DND} lists the performance of all competing methods
on the DND benchmark\footnote{https://noise.visinf.tu-darmstadt.de/}.
From the table, it is easy to be seen that our proposed VDN surpasses all the competing methods.
It is worth noting that CBDNet has the same optimized network with us, containing a \textit{S-Net} designed
for estimating the noise distribution and a \textit{D-Net} for denoising. The superiority of VDN compared
with CBDNet mainly benefits from the deep variational inference optimization.

For easy visualization, on one typical denoising example, results of the best four competing methods are displayed
in Fig.~\ref{fig:SIDD_validation}. Obviously, WNNM is ubable to remove the complex real noise, maybe because
the low-rankness prior is insufficient to describe all the image information and the IID Gaussian noise
assumption is in conflict with the real noise. With the powerful feature extraction ability of CNN, DnCNN and
CBDNet obtain much better denoising results than WNNM, but still with a little noise.
However, the denoising result of 
our proposed VDN has almost no noise and is very close to the groundtruth.

\begin{figure}[t]
    \centering\vspace{-1mm}
    \includegraphics[scale=0.51]{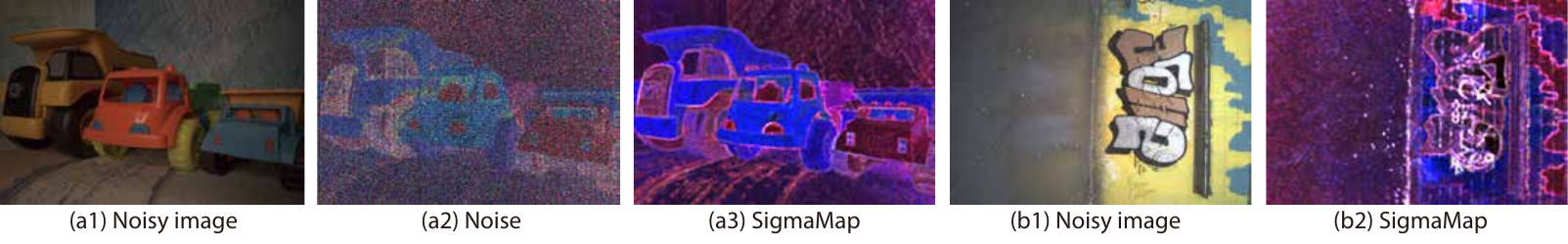}
    \vspace{-2mm}
    \caption{\small{The noise variance map predicted by our proposed VDN on SIDD and DND benchmarks. (a1-a3): The noisy image, real noise
    ($|\bm{y}-\bm{x}|$) and noise variance map of one typical image of SIDD validation dataset. (b1-b2): The noisy image and predicted
    noise variance map of one typical image of DND dataset.}}
    \label{fig:sigmamap_benchmark}
\end{figure}

In Fig.~\ref{fig:sigmamap_benchmark}, we displayed the noise variance map predicted by \textit{S-Net} on the two real
benchmarks. The variance maps had been enlarged several times for easy visualization. It is easy to see that the
predicted noise variance map relates to the image content,
which is consistent with the well-known signal-depend
property of real noise to some extent.

\begin{table}[t]
    \parbox[b]{.55\textwidth}{
    \raggedright
    \caption{\scriptsize Performance of VDN under different $\varepsilon^2_0$ values on SIDD validation
    dataset ($p=7$).} \label{tab:eps_comparison}
    \scriptsize \vspace{-2mm}
    \begin{tabular}{C{0.75cm}|C{0.65cm}C{0.65cm}C{0.65cm}C{0.65cm}C{0.65cm}C{0.75cm}}
        \Xhline{0.8pt}
        $\varepsilon_0^2$  &1e-4     &1e-5   &1e-6            &1e-7    &1e-8      &MSE     \\
        \Xhline{0.4pt}
        PSNR               &38.89    &39.20  &\textbf{39.28}  &39.05   &39.03     &39.01   \\
        \Xhline{0.4pt}
        SSIM               &0.9046   &0.9079 &\textbf{0.9086} &0.9064  &0.9063     &0.9061  \\
        \Xhline{0.8pt}
    \end{tabular}} \hspace{0.2cm} 
    \parbox[b]{.43\textwidth}{
    \raggedleft
    \caption{\scriptsize Performance of VDN under different $p$ values on SIDD validation
    dataset ($\varepsilon_0^2=1e\text{-}6$).} \label{tab:hyper_p}
    \scriptsize \vspace{-2mm}
    \begin{tabular}{C{0.7cm}|C{0.55cm}C{0.55cm}C{0.55cm}C{0.55cm}C{0.55cm}}
        \Xhline{0.8pt}
        $p$                &5                &7                         &11               &15        &19      \\
        \Xhline{0.4pt}
        PSNR               &39.26            &\textbf{39.28}            &39.26            &39.24     &39.24       \\
        \Xhline{0.4pt}             
        SSIM               &\textbf{0.9089}  &0.9086                    &0.9086           &0.9079    &0.9079       \\
        \Xhline{0.8pt}
    \end{tabular}}
    \vspace{-3mm}
\end{table}
\vspace{-1mm}\subsection{Hyper-parameters Analysis}\vspace{-1mm}
The hyper-parameter $\varepsilon_0$ in Eq.~\eqref{Eq:prior-z} determines how much does the desired latent clean
image $\bm{z}$ depend on the simulated groundtruth $\bm{x}$. As discussed in Section~\ref{sec:discussion}, the
negative variational lower bound degenerates into MSE loss when $\varepsilon_0$ is setted as an extremely small
value close to 0. The performance of VDN under different $\varepsilon_0$ values on the SIDD validation dataset
is listed in Table~\ref{tab:eps_comparison}. For explicit comparison, we also directly trained the \textit{D-Net} 
under MSE loss as baseline. From Table~\ref{tab:eps_comparison}, we can see that: 1) when $\varepsilon_0$ is too
large, the proposed VDN obtains relatively worse results since the prior constraint on $\bm{z}$ by simulated
groundtruth $\bm{x}$ becomes weak; 2) with $\varepsilon_0$ decreasing, the performance of VDN tends to be similar
with MSE loss as analysised in theory; 3) the results of VDN surpasses MSE loss about 0.3 dB PSNR
when $\varepsilon_0^2=1e\text{-}6$, which verifies the importantance of noise modeling in our method. Therefore, 
we suggest that the $\varepsilon_0^2$ is set as $1e\text{-}5$ or $1e\text{-}6$ in the real-world denoising task.

In Eq.~\eqref{Eq:prior-sigma}, we introduced a conjugate inverse gamma distribution as prior for $\bm{\sigma}^2$.
The mode of this inverse gamma distribution $\xi_i$ provides a rational approximate evaluation for $\sigma_i^2$,
which is a local estimation in a $p \times p$ window centered at the $i^{th}$ pixel. We compared the performance
of VDN under different $p$ values on the SIDD validation dataset in Table~\ref{tab:hyper_p}. Empirically, VDN
performs consistently well for the hyper-parameter $p$.

\vspace{-2mm}\section{Conclusion}\vspace{-2mm}
We have proposed a new variational inference algorithm, namely varitional denoising network (VDN),
for blind image denoising. The main idea is to learn an approximate posterior to the true posterior
with the latent variables (including clean image and noise variances) conditioned on the input noisy image.
Using this variational posterior expression, both tasks of blind image denoising and noise estimation can
be naturally attained in a unique Bayesian framework. The proposed VDN is a generative method, which can easily
estimate the noise distribution from the input data. Comprehensive experiments have demonstrated the superiority
of VDN to previous works on blind image denoising. Our method can also facilitate the study of other low-level
vision tasks, such as super-resolution and deblurring. Specifically, the fidelity term in these tasks can be
more faithfully set under the estimated non-i.i.d. noise distribution by VDN, instead of the traditional
i.i.d. Gaussian noise assumption.

\vspace{1mm}
\noindent\textbf{Acknowledgements}:This research was supported by National Key R\&D Program of China (2018YFB1004300), the China NSFC project
under contract 61661166011, 11690011, 61603292, 61721002 and U1811461, and Kong Kong RGC General Research Fund
(PolyU 152216/18E).

\bibliography{nips_reference}

\begin{thebibliography}{10}

\bibitem{Abdelhamed_2018_CVPR}
Abdelrahman Abdelhamed, Stephen Lin, and Michael~S. Brown.
\newblock A high-quality denoising dataset for smartphone cameras.
\newblock In {\em The IEEE Conference on Computer Vision and Pattern
  Recognition (CVPR)}, June 2018.

\bibitem{agostinelli2013adaptive}
Forest Agostinelli, Michael~R Anderson, and Honglak Lee.
\newblock Adaptive multi-column deep neural networks with application to robust
  image denoising.
\newblock In {\em Advances in Neural Information Processing Systems}, pages
  1493--1501, 2013.

\bibitem{aharon2006k}
Michal Aharon, Michael Elad, Alfred Bruckstein, et~al.
\newblock K-svd: An algorithm for designing overcomplete dictionaries for
  sparse representation.
\newblock {\em IEEE Transactions on signal processing}, 54(11):4311, 2006.

\bibitem{anaya2014renoir}
Josue Anaya and Adrian Barbu.
\newblock Renoir - a dataset for real low-light noise image reduction.
\newblock {\em arXiv preprint arXiv:1409.8230}, 2014.

\bibitem{amfm_pami2011}
Pablo Arbelaez, Michael Maire, Charless Fowlkes, and Jitendra Malik.
\newblock Contour detection and hierarchical image segmentation.
\newblock {\em IEEE Trans. Pattern Anal. Mach. Intell.}, 33(5):898--916, May
  2011.

\bibitem{bishop2006pattern}
Christopher~M Bishop.
\newblock {\em Pattern recognition and machine learning}.
\newblock springer, 2006.

\bibitem{blei2006variational}
David~M Blei, Michael~I Jordan, et~al.
\newblock Variational inference for dirichlet process mixtures.
\newblock {\em Bayesian analysis}, 1(1):121--143, 2006.

\bibitem{brooks2018unprocessing}
Tim Brooks, Ben Mildenhall, Tianfan Xue, Jiawen Chen, Dillon Sharlet, and
  Jonathan~T Barron.
\newblock Unprocessing images for learned raw denoising.
\newblock {\em arXiv preprint arXiv:1811.11127}, 2018.

\bibitem{buades2005non}
Antoni Buades, Bartomeu Coll, and J-M Morel.
\newblock A non-local algorithm for image denoising.
\newblock In {\em 2005 IEEE Computer Society Conference on Computer Vision and
  Pattern Recognition (CVPR'05)}, volume~2, pages 60--65. IEEE, 2005.

\bibitem{burger2012image}
Harold~C Burger, Christian~J Schuler, and Stefan Harmeling.
\newblock Image denoising: Can plain neural networks compete with bm3d?
\newblock In {\em 2012 IEEE conference on computer vision and pattern
  recognition}, pages 2392--2399. IEEE, 2012.

\bibitem{4271520}
K.~{Dabov}, A.~{Foi}, V.~{Katkovnik}, and K.~{Egiazarian}.
\newblock Image denoising by sparse 3-d transform-domain collaborative
  filtering.
\newblock {\em IEEE Transactions on Image Processing}, 16(8):2080--2095, Aug
  2007.

\bibitem{DengCHI14}
Jia Deng, Olga Russakovsky, Jonathan Krause, Michael Bernstein, Alexander~C.
  Berg, and Li~Fei-Fei.
\newblock Scalable multi-label annotation.
\newblock In {\em ACM Conference on Human Factors in Computing Systems (CHI)},
  2014.

\bibitem{dong2013nonlocal}
Weisheng Dong, Guangming Shi, and Xin Li.
\newblock Nonlocal image restoration with bilateral variance estimation: a
  low-rank approach.
\newblock {\em IEEE transactions on image processing}, 22(2):700--711, 2013.

\bibitem{dong2012nonlocally}
Weisheng Dong, Lei Zhang, Guangming Shi, and Xin Li.
\newblock Nonlocally centralized sparse representation for image restoration.
\newblock {\em IEEE transactions on Image Processing}, 22(4):1620--1630, 2012.

\bibitem{goodfellow2016deep}
Ian Goodfellow, Yoshua Bengio, and Aaron Courville.
\newblock {\em Deep learning}.
\newblock MIT press, 2016.

\bibitem{gu2014weighted}
Shuhang Gu, Lei Zhang, Wangmeng Zuo, and Xiangchu Feng.
\newblock Weighted nuclear norm minimization with application to image
  denoising.
\newblock In {\em Proceedings of the IEEE conference on computer vision and
  pattern recognition}, pages 2862--2869, 2014.

\bibitem{guo2018toward}
Shi Guo, Zifei Yan, Kai Zhang, Wangmeng Zuo, and Lei Zhang.
\newblock Toward convolutional blind denoising of real photographs.
\newblock {\em arXiv preprint arXiv:1807.04686}, 2018.

\bibitem{he2015delving}
Kaiming He, Xiangyu Zhang, Shaoqing Ren, and Jian Sun.
\newblock Delving deep into rectifiers: Surpassing human-level performance on
  imagenet classification.
\newblock In {\em Proceedings of the IEEE international conference on computer
  vision}, pages 1026--1034, 2015.

\bibitem{jain2009natural}
Viren Jain and Sebastian Seung.
\newblock Natural image denoising with convolutional networks.
\newblock In {\em Advances in neural information processing systems}, pages
  769--776, 2009.

\bibitem{kim2016accurate}
Jiwon Kim, Jung Kwon~Lee, and Kyoung Mu~Lee.
\newblock Accurate image super-resolution using very deep convolutional
  networks.
\newblock In {\em Proceedings of the IEEE conference on computer vision and
  pattern recognition}, pages 1646--1654, 2016.

\bibitem{Kingma2015}
Diederik~P. {Kingma} and Jimmy~Lei {Ba}.
\newblock Adam: A method for stochastic optimization.
\newblock {\em international conference on learning representations}, 2015.

\bibitem{kingma2013auto}
Diederik~P Kingma and Max Welling.
\newblock Auto-encoding variational bayes.
\newblock {\em arXiv preprint arXiv:1312.6114}, 2013.

\bibitem{lebrun2015multiscale}
Marc Lebrun, Miguel Colom, and Jean-Michel Morel.
\newblock Multiscale image blind denoising.
\newblock {\em IEEE Transactions on Image Processing}, 24(10):3149--3161, 2015.

\bibitem{lefkimmiatis2018universal}
Stamatios Lefkimmiatis.
\newblock Universal denoising networks: a novel cnn architecture for image
  denoising.
\newblock In {\em Proceedings of the IEEE Conference on Computer Vision and
  Pattern Recognition}, pages 3204--3213, 2018.

\bibitem{liu2018non}
Ding Liu, Bihan Wen, Yuchen Fan, Chen~Change Loy, and Thomas~S Huang.
\newblock Non-local recurrent network for image restoration.
\newblock In {\em Advances in Neural Information Processing Systems}, pages
  1673--1682, 2018.

\bibitem{ma2017waterloo}
Kede Ma, Zhengfang Duanmu, Qingbo Wu, Zhou Wang, Hongwei Yong, Hongliang Li,
  and Lei Zhang.
\newblock {Waterloo Exploration Database}: New challenges for image quality
  assessment models.
\newblock {\em IEEE Transactions on Image Processing}, 26(2):1004--1016, Feb.
  2017.

\bibitem{maggioni2013nonlocal}
Matteo Maggioni, Vladimir Katkovnik, Karen Egiazarian, and Alessandro Foi.
\newblock Nonlocal transform-domain filter for volumetric data denoising and
  reconstruction.
\newblock {\em IEEE transactions on image processing}, 22(1):119--133, 2013.

\bibitem{mairal2008sparse}
Julien Mairal, Michael Elad, and Guillermo Sapiro.
\newblock Sparse representation for color image restoration.
\newblock {\em IEEE Transactions on image processing}, 17(1):53--69, 2008.

\bibitem{mao2016image}
Xiaojiao Mao, Chunhua Shen, and Yu-Bin Yang.
\newblock Image restoration using very deep convolutional encoder-decoder
  networks with symmetric skip connections.
\newblock In {\em Advances in neural information processing systems}, pages
  2802--2810, 2016.

\bibitem{perona1990scale}
Pietro Perona and Jitendra Malik.
\newblock Scale-space and edge detection using anisotropic diffusion.
\newblock {\em IEEE Transactions on pattern analysis and machine intelligence},
  12(7):629--639, 1990.

\bibitem{plotz2017benchmarking}
Tobias Plotz and Stefan Roth.
\newblock Benchmarking denoising algorithms with real photographs.
\newblock In {\em Proceedings of the IEEE Conference on Computer Vision and
  Pattern Recognition}, pages 1586--1595, 2017.

\bibitem{plotz2018neural}
Tobias Pl{\"o}tz and Stefan Roth.
\newblock Neural nearest neighbors networks.
\newblock In {\em Advances in Neural Information Processing Systems}, pages
  1087--1098, 2018.

\bibitem{Ploetz2018}
Tobias Pl\"{o}tz and Stefan Roth.
\newblock Neural nearest neighbors networks.
\newblock In S.~Bengio, H.~Wallach, H.~Larochelle, K.~Grauman, N.~Cesa-Bianchi,
  and R.~Garnett, editors, {\em Advances in Neural Information Processing
  Systems 31}, pages 1087--1098. Curran Associates, Inc., 2018.

\bibitem{ronneberger2015u}
Olaf Ronneberger, Philipp Fischer, and Thomas Brox.
\newblock U-net: Convolutional networks for biomedical image segmentation.
\newblock In {\em International Conference on Medical image computing and
  computer-assisted intervention}, pages 234--241. Springer, 2015.

\bibitem{roth2009fields}
Stefan Roth and Michael~J Black.
\newblock Fields of experts.
\newblock {\em International Journal of Computer Vision}, 82(2):205, 2009.

\bibitem{rudin1992nonlinear}
Leonid~I Rudin, Stanley Osher, and Emad Fatemi.
\newblock Nonlinear total variation based noise removal algorithms.
\newblock {\em Physica D: nonlinear phenomena}, 60(1-4):259--268, 1992.

\bibitem{simoncelli1996noise}
Eero~P Simoncelli and Edward~H Adelson.
\newblock Noise removal via bayesian wavelet coring.
\newblock In {\em Proceedings of 3rd IEEE International Conference on Image
  Processing}, volume~1, pages 379--382. IEEE, 1996.

\bibitem{tai2017memnet}
Ying Tai, Jian Yang, Xiaoming Liu, and Chunyan Xu.
\newblock Memnet: A persistent memory network for image restoration.
\newblock In {\em Proceedings of the IEEE international conference on computer
  vision}, pages 4539--4547, 2017.

\bibitem{tsin2001statistical}
Yanghai Tsin, Visvanathan Ramesh, and Takeo Kanade.
\newblock Statistical calibration of ccd imaging process.
\newblock In {\em Proceedings Eighth IEEE International Conference on Computer
  Vision. ICCV 2001}, volume~1, pages 480--487. IEEE, 2001.

\bibitem{xie2012image}
Junyuan Xie, Linli Xu, and Enhong Chen.
\newblock Image denoising and inpainting with deep neural networks.
\newblock In {\em Advances in neural information processing systems}, pages
  341--349, 2012.

\bibitem{Xu_2018_ECCV}
Jun Xu, Lei Zhang, and David Zhang.
\newblock A trilateral weighted sparse coding scheme for real-world image
  denoising.
\newblock In {\em The European Conference on Computer Vision (ECCV)}, September
  2018.

\bibitem{yong2017robust}
Hongwei Yong, Deyu Meng, Wangmeng Zuo, and Lei Zhang.
\newblock Robust online matrix factorization for dynamic background
  subtraction.
\newblock {\em IEEE transactions on pattern analysis and machine intelligence},
  40(7):1726--1740, 2017.

\bibitem{yue2018hyperspectral}
Zongsheng Yue, Deyu Meng, Yongqing Sun, and Qian Zhao.
\newblock Hyperspectral image restoration under complex multi-band noises.
\newblock {\em Remote Sensing}, 10(10):1631, 2018.

\bibitem{zhang2017beyond}
Kai Zhang, Wangmeng Zuo, Yunjin Chen, Deyu Meng, and Lei Zhang.
\newblock Beyond a gaussian denoiser: Residual learning of deep cnn for image
  denoising.
\newblock {\em IEEE Transactions on Image Processing}, 26(7):3142--3155, 2017.

\bibitem{zhang2018ffdnet}
Kai Zhang, Wangmeng Zuo, and Lei Zhang.
\newblock Ffdnet: Toward a fast and flexible solution for cnn-based image
  denoising.
\newblock {\em IEEE Transactions on Image Processing}, 27(9):4608--4622, 2018.

\bibitem{zhang2018residual}
Yulun Zhang, Yapeng Tian, Yu~Kong, Bineng Zhong, and Yun Fu.
\newblock Residual dense network for image super-resolution.
\newblock In {\em Proceedings of the IEEE Conference on Computer Vision and
  Pattern Recognition}, pages 2472--2481, 2018.

\bibitem{zhu2017blind}
Fengyuan Zhu, Guangyong Chen, Jianye Hao, and Pheng-Ann Heng.
\newblock Blind image denoising via dependent dirichlet process tree.
\newblock {\em IEEE transactions on pattern analysis and machine intelligence},
  39(8):1518--1531, 2017.

\bibitem{zhu2016noise}
Fengyuan Zhu, Guangyong Chen, and Pheng-Ann Heng.
\newblock From noise modeling to blind image denoising.
\newblock In {\em Proceedings of the IEEE Conference on Computer Vision and
  Pattern Recognition}, pages 420--429, 2016.

\end{thebibliography}
\end{document}